%% file: main.tex
\documentclass[10pt,twocolumn,letterpaper]{article}

\usepackage{cvpr}    % To produce the REVIEW version
\input{preamble}

\definecolor{cvprblue}{rgb}{0.21,0.49,0.74}
\usepackage[pagebackref,breaklinks,colorlinks,allcolors=cvprblue]{hyperref}

\def\paperID{39007} % *** Enter the Paper ID here
\def\confName{CVPR}
\def\confYear{2026} 

\title{Context-Nav: Context-Driven Exploration and Viewpoint-Aware 3D Spatial Reasoning for Instance Navigation}

\author{Won Shik Jang\textsuperscript{\rm 1} and Ue-Hwan Kim\textsuperscript{\rm 1}\thanks{Corresponding author}\\
\textsuperscript{\rm 1}Department of AI Convergence\\
Gwangju Institute of Science and Technology, Gwangju, South Korea \\
{\tt\small wonsicjang@gm.gist.ac.kr, uehwan@gist.ac.kr}
}

\begin{document}
%\maketitle

\twocolumn[%
    \vspace{-1.3cm}
    \maketitle
    \vspace{-0.5cm}
    % Begin centered environment for the table
    \begin{center}
        \footnotesize
        \setlength{\tabcolsep}{1pt} % Adjust the space between columns if necessary
        \renewcommand{\arraystretch}{1.2} % Adjust the row height if necessary

        %Begin the table
        \begin{tabular}{>{\centering\arraybackslash}m{17cm}}
            {\includegraphics[width=0.95\linewidth]{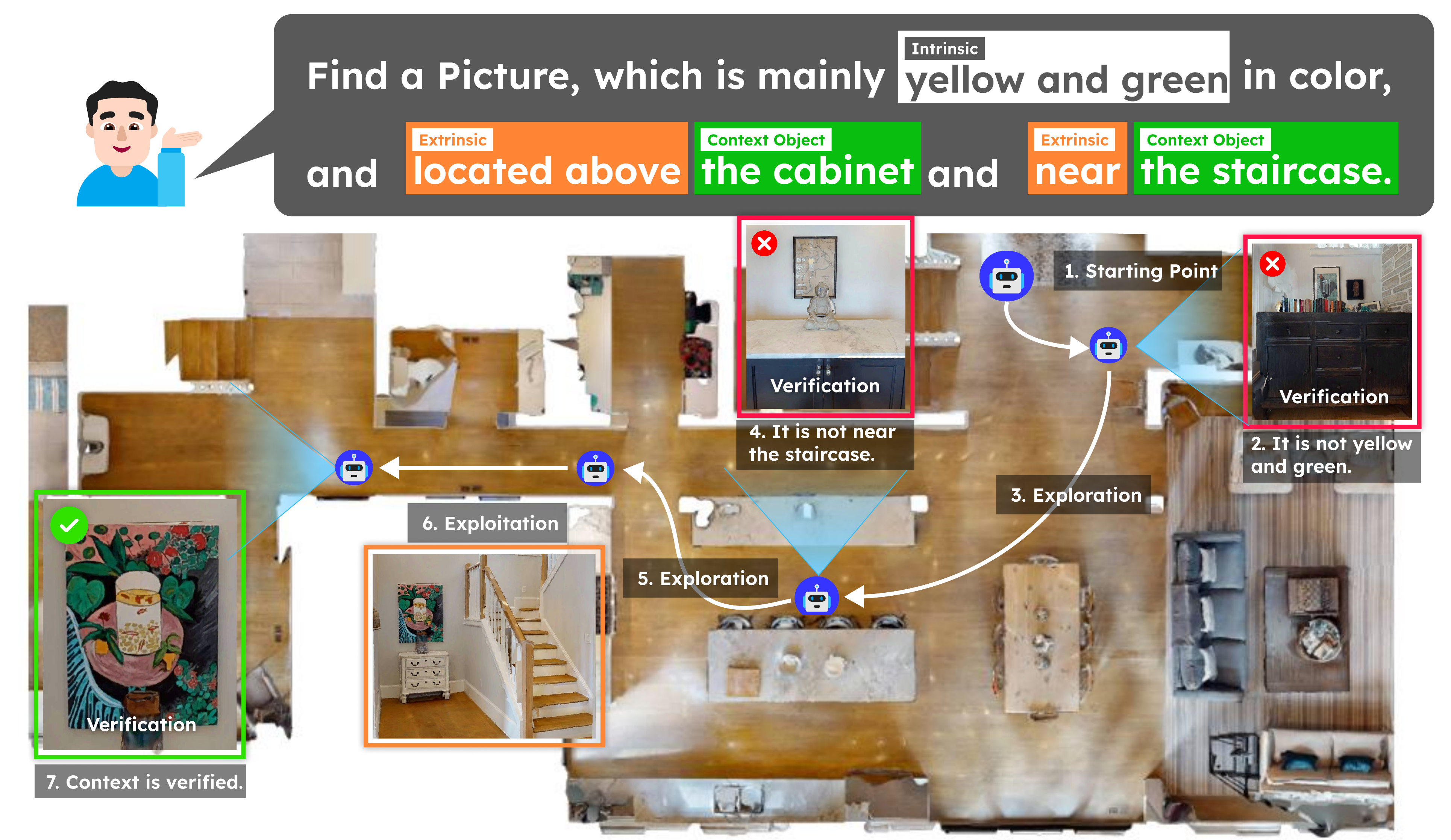}}
        \end{tabular}
        \captionsetup{hypcap=false}
        \captionof{figure}{\textbf{Overview of the text-goal instance navigation task and our context-driven pipeline.} Given a long description that mixes intrinsic attributes (“mainly yellow and green”) with extrinsic context (“located above the cabinet and near the staircase”), the agent explores guided by the context-driven value map and performs viewpoint-aware 3D spatial reasoning. The agent rejects early picture candidates because either the color or nearby context objects do not match, and ultimately exploits the region containing both the cabinet and staircase, where 3D verification confirms that all intrinsic and extrinsic constraints are satisfied.}
        
    \end{center}
    \vspace{0.2cm}
]
\input{bodies/00_Abstract}

\input{bodies/01_introduction}

\input{bodies/02_related_work}
\input{bodies/03_method}

\input{bodies/04_experiments}

\input{bodies/05_conclusion}

\newpage
\input{bodies/06_Acknowledgement}
{
    \small
    \bibliographystyle{ieeenat_fullname}
    \bibliography{main}

}


\end{document}

\renewcommand{\thefigure}{S\arabic{figure}}
\renewcommand{\thetable}{S\arabic{table}}
\clearpage
\appendix
\setcounter{page}{1}
\maketitlesupplementary

\section{Overview}
In this supplementary material, we provide additional details and analyses that complement the main paper:
\begin{itemize}
    \item \textbf{Sec.~B (Related Work)} summarizes additional background on visual navigation goal formulations and zero-shot ObjectNav pipelines.
    \item \textbf{Sec.~C (Text-Goal Preprocessing)} details our LLM-based pipeline that decomposes free-form text goals into intrinsic attributes, context objects, and spatial-relation triples used for downstream reasoning.
    \item \textbf{Sec.~D (Viewpoint-Aware 3D Verification of Extrinsic Attributes)} explains how viewpoint-aware 3D spatial reasoning is performed to verify extrinsic contextual relations between the target and surrounding objects.
    \item \textbf{Sec.~E (Prompts)} lists the exact LLM and VLM prompts used for attribute extraction, question generation, context filtering, relation extraction, non-COCO instance verification, and intrinsic attribute scoring.
    \item \textbf{Sec.~F (Additional Experimental Results)} reports further quantitative results, including a re-evaluation of InstanceNav under stricter success criteria and an analysis of Context-Nav’s performance.
    \item \textbf{Sec.~G (Implementation Details)} provides implementation specifics such as model choices, agent configuration, and hyperparameters for detection, mapping, and wall reconstruction, and additionally reports runtime/latency statistics.
    \item \textbf{Sec.~H (Category Coverage on CoIN-Bench)} summarizes the target category coverage of CoIN-Bench and the resulting diversity of the evaluated open-vocabulary category space.
    \item \textbf{Sec.~I (Failure Case Analysis)} analyzes failure episodes (time-out or false-positive stop) and discusses dominant failure modes, including planning under imperfect geometry, perception/detection failures, intrinsic ambiguity, and room-segmentation errors.
    \item \textbf{Sec.~J (Qualitative Comparison with RL-Trained and Training-Free Baselines)} presents qualitative visualizations comparing Context-Nav with RL-trained and training-free baselines on InstanceNav and CoIN-Bench.
\end{itemize}

\section{Related Work}
Goal formulations in visual navigation differ primarily in how the target is specified and how strongly the trajectory is constrained. For completeness, we expand on the task variants summarized in the main paper. Point-goal navigation \cite{Batra2020ObjectNavRO} specifies the target as a metric coordinate in the environment, and the agent must reach this location using onboard sensing and odometry. Image-goal navigation instead provides a reference image that the agent must match by exploring the scene \cite{zhu2016targetdrivenvisualnavigationindoor, Sun2023FGPromptFG}. Category-level ObjectNav asks the agent to reach any instance of a named class (e.g., ``a chair'') \cite{Batra2020ObjectNavRO, habitatchallenge2023}, while vision-and-language navigation extends this to stepwise natural-language instructions or route descriptions that constrain the path the agent should follow \cite{anderson2018visionandlanguagenavigationinterpretingvisuallygrounded, qi2020reverieremoteembodiedvisual}. These formulations differ in how precisely they specify the final instance and path, and each introduces characteristic limitations: category-level goals under-specify which instance should be selected, image-goal settings presuppose access to an ideal goal snapshot, and instruction-following tasks assume detailed human-authored guidance at test time.

Zero-shot ObjectNav methods relax the fixed task taxonomy by enabling open-vocabulary goal specifications at inference time. A common strategy replaces static category embeddings with CLIP-style text--image encoders or retrieval modules that map free-form textual queries to visual exemplars \cite{majumdar2023zsonzeroshotobjectgoalnavigation, alhalah2022zeroexperiencerequiredplug, sun2024prioritizedsemanticlearningzeroshot}. Training-free modular pipelines further use vision--language cues as an exploration prior: frontier candidates on a top-down occupancy map are ranked using CLIP similarities \cite{gadre2022cowspasturebaselinesbenchmarks} or text-conditioned value maps \cite{yokoyama2023vlfmvisionlanguagefrontiermaps, taioli2025coin}, and large language or vision--language models score these frontiers based on detected-object lists or semantic maps before a classical planner executes the selected action \cite{zhou2023escexplorationsoftcommonsense, Yu_2023, wu2024voronavvoronoibasedzeroshotobject, kuang2024openfmnavopensetzeroshotobject, zhang2024trihelperzeroshotobjectnavigation, yin2024sgnavonline3dscene, cai2023bridgingzeroshotobjectnavigation, long2024instructnavzeroshotgenericinstruction}. While these approaches already exploit language as a high-level prior over frontiers, they typically operate at the category level or with short attribute snippets, and do not yet elevate long, contextual descriptions to the central exploration and verification signal that Context-Nav provides.

\section{Text-Goal Preprocessing}
\label{sec:TG_processing}

\subsection{Benchmark-Specific Goal Decomposition}
Text-goal decomposition follows a shared pipeline across benchmarks, with minor task-specific differences for InstanceNav~\cite{sun2024prioritizedsemanticlearningzeroshot} and CoIN-Bench~\cite{taioli2025coin}. On InstanceNav, the benchmark already provides a target category together with two separate strings: an intrinsic-attribute description and an extrinsic/context description. These two strings are directly treated as the intrinsic part and the context part of $G$. On CoIN-Bench, the benchmark instead provides a single caption that mixes intrinsic and extrinsic cues; this full caption is used both as the input to intrinsic-attribute extraction and as the context-description text for context objects and relations. In both cases, the target category label is supplied to all prompts as metadata.

\subsection{Intrinsic Attributes and Attribute Questions}
Intrinsic attributes of the target are extracted by a single LLM call conditioned on the linguistic description and the target category. Using the intrinsic-attribute extraction prompt (Sec.~\ref{sec:prompts}), the model is instructed to return only attributes that explicitly modify the target category in $G$, restricted to color and shape; attributes of other objects in the caption are ignored. The output is a small key--value dictionary (e.g., \texttt{color: "yellow and green"}, \texttt{shape: "square"}).

The resulting attribute dictionary is then converted into a set of binary questions used for yes/unknown/no intrinsic verification. Through the attribute question generation prompt, the LLM receives the target category and the extracted attributes and returns a set of concise, unambiguous yes/no questions for each attribute type (e.g., ``Is the outlined picture mainly yellow and green?"). These questions are later issued to the VLM when a candidate instance is observed during the intrinsic-verification stage in Sec.~3.4 of the main paper.

\subsection{Context Objects and Spatial Relations}
Context-object categories are derived by filtering and canonicalizing noun phrases extracted from the context part of $G$. First, candidate noun phrases are extracted with a syntactic parser~\cite{Honnibal2015AnIN} and lightly normalized (e.g., ``the wooden cabinet"). Second, the normalized list is passed to the context object filtering prompt, which retains only phrases that refer to concrete physical indoor objects and discards directions, regions, and purely material/color/shape expressions. Third, using the context synonym grouping prompt, the LLM groups the remaining phrases into synonym clusters, assigns a canonical label to each group (aligned with common indoor categories, e.g., ``cabinet"), and enforces that any group containing the target category uses the exact target string as its canonical key. This procedure yields a set of canonical context categories together with a mapping from raw phrases to canonical labels that is reused for relation extraction.

Context descriptions are subsequently converted into symbolic spatial relations over the allowed object terms. Using the spatial relation extraction prompt, the LLM receives the full caption and the list of allowed raw terms (all canonical context categories plus the target category) and is instructed to output a small set of unambiguous pairwise relations $(\text{ref}, \text{tgt}, \rho)$, where $\rho \in \{\texttt{left}, \texttt{right}, \texttt{front}, \texttt{behind}, \texttt{near}, \texttt{above}, \texttt{below}\}$. The prompt constrains the model to use only allowed surface forms for objects and to skip ambiguous pairs; the raw object names are then mapped back to canonical categories. The resulting relation set $\mathcal{T}$ and the associated instance centers form the input to the viewpoint-aware 3D spatial reasoning stage in Sec.~3.4 of the main paper.

\section{Viewpoint-Aware 3D Verification of Extrinsic Attributes}
\begin{figure*}
\centering
\includegraphics[height=0.93\textheight, keepaspectratio]
{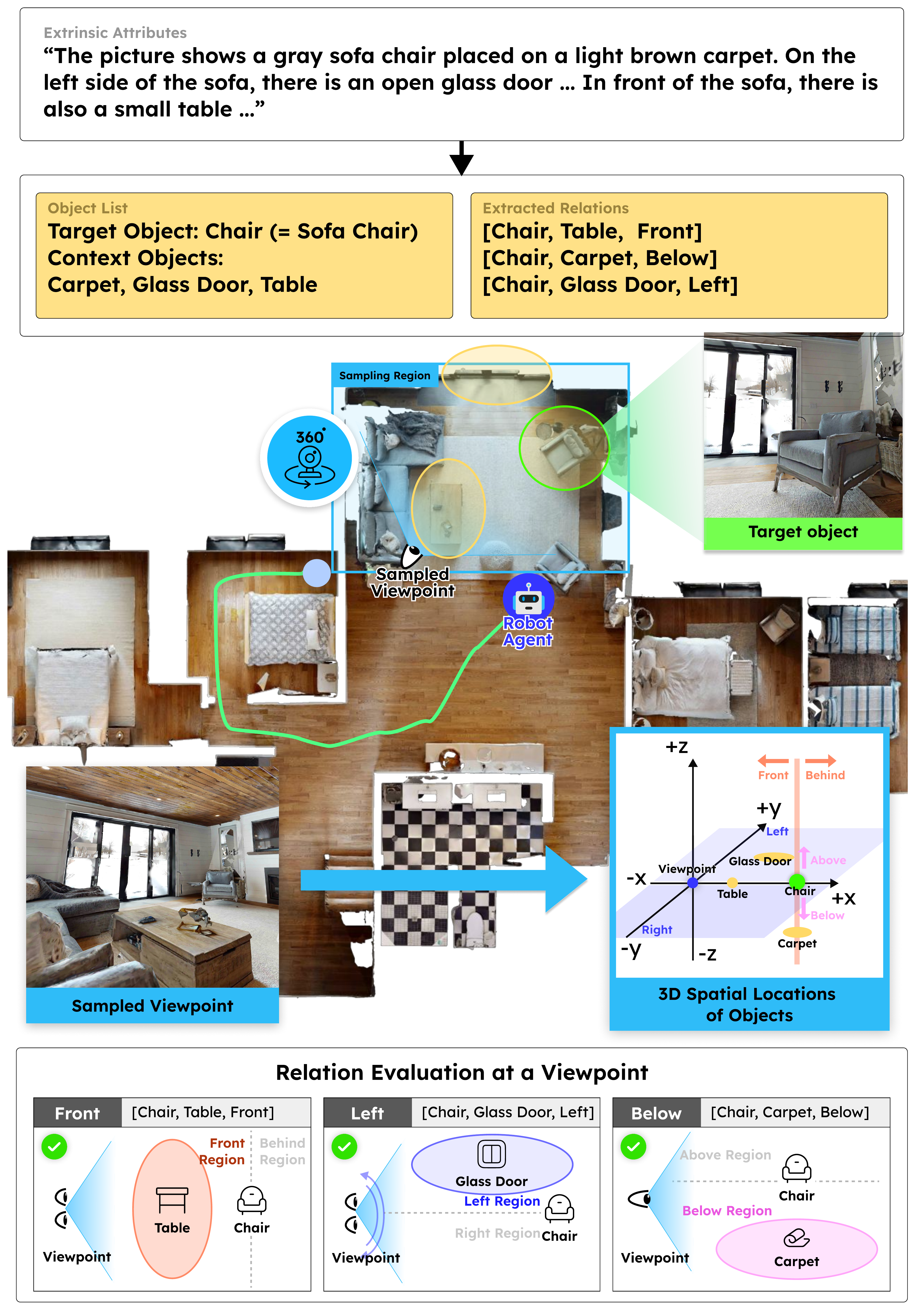}
\caption{\textbf{Viewpoint-aware 3D verification of extrinsic attributes.} Starting from the extrinsic part of the goal description, Context-Nav extracts context objects and spatial-relation triples, builds instance-level 3D point clouds, and samples candidate viewpoints around the reference--target pairs. For each candidate viewpoint, a local frame is aligned and the corresponding spatial predicates are evaluated; the figure visualizes an example viewpoint from which all extrinsic relations are satisfied simultaneously, together with the resulting relation checks for the confirmed target instance.}
\label{fig:relation_evaluation}
\end{figure*}

Viewpoint-aware 3D verification operationalizes extrinsic attributes by testing spatial-relation triples against the reconstructed instance-level map under viewpoint uncertainty. Given the goal $G$, the extrinsic/context part of the description is parsed into spatial-relation triples $(\text{ref}, \text{tgt}, \rho)$, and candidate context instances and relation sets are derived as in Sec.~\ref{sec:TG_processing}. As the agent explores, the perception and mapping modules build instance-level 3D point clouds and a wall-only map that defines wall-bounded rooms. After room-level filtering ensures that the candidate target instance and at least one context instance are co-located in the same wall-bounded room (Step~1 in Sec.~3.4 of the main paper), the pipeline samples the candidate viewpoint set $\mathcal{V}$ around the reference--target pairs, aligns a local frame at each viewpoint, and evaluates the spatial predicates defined in Eq.~(5) of the main paper. Figure~\ref{fig:relation_evaluation} provides a visual overview of this procedure.

\section{Prompts}
\label{sec:prompts}
This section lists the LLM and VLM instructions used in the text-goal preprocessing and verification pipeline. In all prompts, tokens written in \texttt{UPPERCASE} denote placeholders that are replaced at runtime by episode-specific values (e.g., the target category or the goal caption).

\subsection{Intrinsic Attribute Extraction}
In the intrinsic-attribute extraction prompt, \texttt{\{TARGET\}} is replaced with the target category (e.g., ``picture'') and \texttt{\{TEXT\}} with either the intrinsic-attribute string (InstanceNav) or the full caption (CoIN-Bench). The model is expected to return a JSON-style dictionary whose keys are a subset of \texttt{"color"} and \texttt{"shape"} and whose values are free-form textual descriptions.

\begin{tcolorbox}[
    colback=gray!10,
    colframe=gray!,
    boxrule=0pt,
    left=2mm,right=2mm,top=1mm,bottom=1mm
]
You extract visual attributes from a short object description.

INPUT: a TARGET category \{TARGET\} and a description \{TEXT\}.

1. Read the description and identify attributes that explicitly modify the TARGET object.

2. Only consider the following attribute types: color and shape.

3. Ignore attributes of any other object, even if they are in the description.

4. Do not infer or guess; only use attributes that are clearly stated.

OUTPUT: a JSON dictionary with up to two keys:

\{
  ``color": string or null,
  ``shape": string or null
\}

Omit any key whose attribute is not explicitly mentioned.
\end{tcolorbox}

\subsection{Attribute Question Generation}
For attribute question generation, \texttt{\{TARGET\}} is replaced with the target category and \texttt{\{ATTRIBUTES\}} with the JSON dictionary returned by the previous prompt. The model outputs a list of question--attribute-type pairs.

\begin{tcolorbox}[
    colback=gray!10,
    colframe=gray!,
    boxrule=0pt,
    left=2mm,right=2mm,top=1mm,bottom=1mm
]
You convert attributes into direct yes/no VQA questions for a vision-language model.

INPUT: \\
- TARGET category: \{TARGET\} \\
- Attribute dictionary: \{ATTRIBUTES\}, a JSON with keys from [``color", ``shape"].

1. For each attribute present in \{ATTRIBUTES\}, write concise, unambiguous yes/no questions about a single instance of the TARGET.

2. Each question must explicitly mention the TARGET and the attribute value (e.g., color phrase).

3. Do not create questions for attributes that are missing from the dictionary.

4. Do not ask about other objects or unstated properties.

OUTPUT: a JSON list of objects of the form

[
  \{``atype": ``color", ``q": ``...?"\},
  \{``atype": ``color", ``q": ``...?"\},
  \{``atype": ``shape", ``q": ``...?"\},
  ...
]

where \texttt{atype} is either ``color" or ``shape".
\end{tcolorbox}

\subsection{Context Object Filtering}
For context object filtering, \texttt{\{PHRASES\}} is replaced with the list of normalized noun phrases extracted from the context description (one phrase per line or as a JSON list).

\begin{tcolorbox}[
    colback=gray!10,
    colframe=gray!,
    boxrule=0pt,
    left=2mm,right=2mm,top=1mm,bottom=1mm
]
You are given candidate noun phrases from an indoor scene description.

INPUT: a list of noun phrases \{PHRASES\}.

1. Decide which phrases refer to concrete physical objects that can plausibly exist indoors (e.g., ``cabinet", ``staircase", ``pillow").

2. EXCLUDE phrases that are:
   - pure directions or regions (e.g., ``left side", ``corner of the room"),
   - purely generic items without a concrete object (e.g., ``stuff", ``some things"),
   - phrases that only name materials, colors, or shapes (e.g., ``wooden", ``red", ``round object" without a category).

3. Do not invent new words or rewrite phrases; you may only select from the input list.

OUTPUT: a JSON list of the selected phrases, preserving their original text.
\end{tcolorbox}

\subsection{Context Synonym Grouping}
For context synonym grouping, \texttt{\{TERMS\}} is replaced with the filtered phrases from the previous step and \texttt{\{TARGET\}} with the target category.

\begin{tcolorbox}[
    colback=gray!10,
    colframe=gray!,
    boxrule=0pt,
    left=2mm,right=2mm,top=1mm,bottom=1mm
]
You group indoor object names into synonym clusters and assign a canonical label to each cluster.

INPUT: \\
- Filtered phrases: \{TERMS\}\\
- Target category: \{TARGET\}

1. Treat each item in \{TERMS\} as a candidate object name.

2. Group terms that refer to the same indoor household object category (e.g., ``sofa" and ``couch").

3. For each group, choose a canonical label: \\
   - Use a simple, lowercase, space-separated form (no underscores). \\
   - If any term in the group exactly matches \{TARGET\}, then set the canonical label of that group to \{TARGET\} exactly.

4. Do not introduce new object names; every term in each group must come from \{TERMS\}.

OUTPUT: a JSON object

\{
  ``canonical\_1": [``term\_a", ``term\_b", ...],
  ``canonical\_2": [``term\_c", ...],
  ...
\}

covering all grouped terms.
\end{tcolorbox}

\newpage
\subsection{Spatial Relation Extraction}
For spatial relation extraction, \texttt{\{CAPTION\}} is replaced with the full goal description $G$ and \texttt{\{ALLOWED\_TERMS\}} with the list of allowed raw object names (target category plus all canonical context terms and their raw variants).

\begin{tcolorbox}[
    enhanced,
    breakable,
    colback=gray!10,
    colframe=gray!,
    boxrule=0pt,
    left=2mm,right=2mm,top=1mm,bottom=1mm
]
You convert a single-view image caption into pairwise spatial relations among indoor objects.

INPUT: \\
- Caption describing an indoor scene: \{CAPTION\} \\
- Allowed object names (surface forms): \{ALLOWED\_TERMS\}

1. Identify all mentions of objects whose names (or synonyms) appear in \{ALLOWED\_TERMS\}.

2. From the caption, extract pairwise spatial relations of the form (REFERENCE object, TARGET object, RELATION), where RELATION is one of: \\
   - ``left", ``right", ``front", ``behind", ``near", ``above", ``below".

3. The caption may describe frame-relative positions (e.g., ``left side of the image", ``upper right corner"). Interpret them in terms of 2D image coordinates only (no depth reasoning), e.g.: \\
   - Object on the left/upper-left/lower-left of the frame is LEFT of an object on the right/upper-right/lower-right. \\
   - Object in the lower/bottom region is BELOW an object in the upper/top region.

4. Use ONLY the provided \{ALLOWED\_TERMS\} as surface forms for ``ref" and ``tgt".
   If the text uses a synonym not in the list, map it to the nearest allowed label and output the allowed label.

5. Emit only relations that are clearly entailed by the caption.
   Skip ambiguous, underspecified, or contradictory relations (e.g., both left(A,B) and left(B,A)).

6. Prefer a concise set of relations: output at most 6 of the most salient pairs.

OUTPUT: a JSON list of relations, each of the form

\{
  {``ref": ``object\_1", ``tgt": ``object\_2", ``rtype": ``relation"},
  ...
\}
where \texttt{rtype} is one of [``left",``right",``front",``behind",``near",``above",``below"].
\end{tcolorbox}

\subsection{Non-COCO Instance Verification}
Open-set categories that are not covered by COCO are verified by prompting the VLM to decide whether a mask-highlighted region belongs to a proposed category. The placeholder \texttt{\{CATEGORY\_NAME\}} is filled with the detector's open-vocabulary label (e.g., ``dresser", ``radiator"), and the RGB frame is annotated with an outline around the candidate region before being passed to the VLM. The returned \texttt{prob} is interpreted as a calibrated confidence in $[0,1]$, and instances are treated as valid when \texttt{prob} $\ge 0.6$.

\begin{tcolorbox}[
    colback=gray!10,
    colframe=gray!,
    boxrule=0pt,
    left=2mm,right=2mm,top=1mm,bottom=1mm
]
Is the outlined object an instance of the category ``\{CATEGORY\_NAME\}"?
Judge ONLY the pixels inside the outline/mask; ignore everything else.
If the outline is faint, assume the highlighted/outlined region marks a SINGLE candidate instance. \\
Return ONLY JSON: \{``prob": float between 0 and 1\} where prob is P(instance $\in$ category).
If the region is too small/blurred/occluded or ambiguous, return a value near 0.5.
\end{tcolorbox}

\subsection{Attribute Scoring for Intrinsic Verification}
For intrinsic attribute verification, the VLM is instructed to act as a strict visual judge that scores whether a single attribute claim about the highlighted instance is true, using only the given image. The placeholder \texttt{\{ATTRIBUTE\_QUESTION\}} is filled with a paraphrased yes/unknown/no-style question produced by the LLM, and the resulting integer score in $\{0,\dots,15\}$ is mapped to the Yes/Unknown/No bins described in Sec.~3.4 of the main paper:

\begin{tcolorbox}[
    colback=gray!10,
    colframe=gray!,
    boxrule=0pt,
    left=2mm,right=2mm,top=1mm,bottom=1mm
]
You are a strict visual judge. Score whether the claim about the object is true using ONLY the provided image.
SCORING (integer 0..15): \\
- NO [0..4] \\
- YES [11..15] \\
- UNKNOWN [5..10] (visibility limits only).\\
Never output text other than the JSON below.
CLAIM: Question about the outlined object: \{ATTRIBUTE\_QUESTION\} \\
Return ONLY valid JSON: \{``score": integer 0..15\}
\end{tcolorbox}

\newpage
\section{Additional Experimental Results}

\begin{table}[h]
\centering
\begin{threeparttable}
\small
\begin{tabular}{l cc cc}
\toprule
\multirow{2}{*}{Method} 
& \multicolumn{2}{c}{Model Condition} 
& \multicolumn{2}{c}{InstanceNav} \\
\cmidrule(lr){2-3}\cmidrule(lr){4-5}
 & Input & Training-free
 & SR$\uparrow$ & SPL$\uparrow$ \\
\midrule
PSL \cite{sun2024prioritizedsemanticlearningzeroshot}  & d & \xmark 
& 18.0 & 7.2 \\
VLFM \cite{yokoyama2023vlfmvisionlanguagefrontiermaps}  & c & \cmark 
& 9.5 & 5.7 \\
UniGoal \cite{yin2025unigoaluniversalzeroshotgoaloriented}  & d & \cmark 
& 13.5 & 6.7 \\
\textbf{Ours} & d & \cmark 
& \textbf{21.2} & \textbf{7.9} \\
\bottomrule
\end{tabular}
\caption{\textbf{Benchmark results on InstanceNav.} Comparison of RL-trained policies, training-free modular baselines, and the proposed Context-Nav on InstanceNav \cite{sun2024prioritizedsemanticlearningzeroshot} under the stricter CoIN-Bench \cite{taioli2025coin} success criteria. Input type c denotes a category-level goal specification, while d denotes a language description of the target.}
\label{table:main_result_metric_variation}
\end{threeparttable}
\end{table}
Additional experiments re-evaluating InstanceNav~\cite{sun2024prioritizedsemanticlearningzeroshot} under the stricter CoIN-Bench~\cite{taioli2025coin} success criteria show that Context-Nav maintains state-of-the-art SR and SPL among both RL-trained and training-free methods. As summarized in Table~\ref{table:main_result_metric_variation}, CoIN-Bench reduces the success radius from $1$m to $0.25$m and shortens the episode horizon from $1{,}000$ to $500$ steps, forcing agents to stop precisely at the described instance among same-category distractors while exploring efficiently. Under this strict setting, Context-Nav achieves 21.2\% SR and 7.9 SPL, indicating that leveraging contextual information with viewpoint-aware 3D reasoning remains effective even when approximate stops and long trajectories are strongly penalized.

\section{Implementation Details}
\subsection{Language and Vision-Language Models}
Language and vision--language components of Context-Nav are instantiated with GPT-OSS 20B~\cite{openai2025gptoss120bgptoss20bmodel} and Qwen2.5-VL 7B~\cite{Qwen2.5-VL}, respectively, for all experiments. GPT-OSS 20B parses the free-form goal $G$ into intrinsic and extrinsic attributes, extracts relation triples $(\text{ref}, \text{tgt}, \rho)$, and generates paraphrased yes/unknown/no question templates for attribute verification (Sec.~\ref{sec:TG_processing}). Qwen2.5-VL 7B is employed both as the open-set category verifier in Sec.~3.2 of the main paper (for non-COCO categories that require yes/no classification on masked regions) and as the intrinsic-attribute VQA module in Sec.~3.4, where it answers attribute-specific questions on the highlighted object and returns discretized confidence scores $s \in \{0,\dots,15\}$. Each model is used with its default inference hyperparameters (temperature, top-$p$), and no fine-tuning or in-domain adaptation is performed.

\subsection{Agent Configuration}
Agent embodiment and action space are kept identical across InstanceNav and CoIN-Bench. In both benchmarks, the agent uses the discrete action set ${\texttt{forward}, \texttt{ turn-left}, \texttt{ turn-right}, \texttt{ stop}}$ as defined in Sec.~3.1 of the main paper. A \texttt{forward} action moves the agent by $0.25$m along its current heading, while \texttt{turn-left} and \texttt{turn-right} rotate the agent by $30^{\circ}$.

\subsection{Hyperparameters}
\noindent\textbf{Detection thresholds.}
Detection thresholds in the open-vocabulary detection and verification module (Sec.~3.2, ``Open-Vocabulary Detection and Verification'') are configured as follows. GroundingDINO~\cite{liu2024groundingdinomarryingdino} proposals are accepted when their confidence is at least $0.45$, YOLOv7~\cite{wang2022yolov7trainablebagoffreebiessets} detections are accepted when the confidence is at least $0.8$, and VLM yes/no classification outputs are promoted to persistent instances when their score in $[0,1]$ is greater than or equal to $0.6$.

\noindent\textbf{Spatial proximity heuristic.}
The spatial proximity heuristic for first-stage association in the instance-level 3D mapping module (Sec.~3.2, ``Instance-Level 3D Mapping'') merges two observations when the Euclidean distance between their 2D centers on the ground plane is below $0.26$m, thereby cheaply capturing short-term revisits.

\noindent\textbf{Voxel-overlap association.}
Voxel-overlap association in Eq.~(1) of the main paper relies on discretizing each instance point cloud into a voxel grid with resolution $0.05$m. Up to $5{,}000$ points are uniformly sampled per instance (or all points if fewer are available), and the overlap score $s(A,B)$ is computed on the resulting voxel sets. Two instances are merged when $s(A,B) > 0.45$; otherwise, a new instance is created.

\noindent\textbf{Wall-map RANSAC.}
RANSAC parameters for constructing the wall-only map (Sec.~3.2, ``Wall-Only Map'') are chosen to robustly recover structural planes. Depth points are range-gated to $[0.5, 5]$ meters before RANSAC~\cite{Fischler1981RANSAC, Zhou2018}; a $0.03$m inlier distance threshold is used, at least $400$ inliers per plane are required, and the maximum number of iterations is capped at $1{,}500$. A vertical-plane constraint on the fitted normal ($n_z \le 0.3$) is additionally enforced, and at most three structural planes per frame are retained when updating the wall-only layer.

\subsection{Runtime and Latency}
\begin{table}[ht]
    \vspace{-5pt}
    \centering
    \resizebox{\linewidth}{!}{
    \begin{tabular}{llccccc}
        \toprule &
        \textbf{Component} &
        \makecell{\textbf{Det/Seg}\\\textbf{Modules}} &
        \makecell{\textbf{Value-map}\\\textbf{Update}} &
        \makecell{\textbf{3D}\\\textbf{Association}} &
        \makecell{\textbf{VLM}\\\textbf{Inference}} &
        \makecell{\textbf{Viewpoint-aware}\\\textbf{Verification}} \\
        \midrule
        & \textbf{Context-Nav} & 0.12 & 0.11 & 0.29 & 0.49 & 0.0004 \\
        \bottomrule
    \end{tabular}}
    \vspace{-9pt}
    \caption{\textbf{Per-call latency (in seconds) of each module.}}
    \label{tab:latency}
    \vspace{-9pt}
\end{table}

On a single NVIDIA A100 GPU, Context-Nav runs at 0.54~s/step on average (range: 0.36--1.12~s/step).
Table~\ref{tab:latency} summarizes the per-call latency of each component.
Crucially, VLM inference is not executed at every step: it is triggered in only 10\% of steps
(8\% for open-set category verification and 2\% for intrinsic-attribute verification), and most steps
incur no VLM overhead.
For reference, representative modular baselines are substantially slower:
AIUTA runs at 1.8~s/step on average (0.9--7.9~s/step), and UniGoal runs at 0.8~s/step (0.4--12~s/step).

\section{Category Coverage on CoIN-Bench}
Category coverage on CoIN-Bench \cite{taioli2025coin} is broader and more diverse than on InstanceNav \cite{sun2024prioritizedsemanticlearningzeroshot} while still overlapping with the six InstanceNav categories (chair, sofa, TV, bed, toilet, and plant). Across both benchmarks, the evaluation therefore includes common InstanceNav targets such as chair, bed, plant, and TV as well as a wide range of additional furniture, storage, textile, and decorative objects, allowing us to test whether context-driven exploration and 3D spatial verification generalize beyond a small, fixed category set.

The CoIN-Bench splits further diversify the target space. On \texttt{Val Seen}, the agent is evaluated on categories including cabinet, bed, table, clothes, kitchen lower cabinet, blanket, cloth, TV, bathroom cabinet, wardrobe, desk, display cabinet, chair, heater, rack, and board. The \texttt{Val Unseen} split introduces additional target categories such as mirror, picture, dresser, radiator, plant, window glass, hanging clothes, rug, and book. Finally, the \texttt{Val Seen Synonyms} split includes kitchen cabinet, armchair, and shelf, and deliberately tests robustness to lexical variation (e.g., armchair vs.\ chair) and fine-grained category distinctions. This combination of overlapping InstanceNav categories and split-specific CoIN-Bench targets ensures that the reported SR and SPL reflect performance across a heterogeneous, open-vocabulary category space rather than a narrow, benchmark-specific label set.

\section{Failure Case Analysis}
We define a failure episode as either (i) time-out, where the agent exceeds the maximum step budget
without reaching the described instance, or (ii) a false-positive stop, where the agent terminates at
a distractor of the same category or an off-target object.
We group failures into three dominant modes.

\noindent\textbf{Planning.} A common failure source is planning under imperfect geometry.
In some scenes, phantom free-space artifacts in HM3D lead the global planner to produce paths that are not executable in practice.
In addition, the agent may fail to perceive obstacles early enough due to its limited field of view, which can cause late replanning and
unrecoverable detours near the end of an episode.

\noindent\textbf{Detection.}
Failures in the perception stack typically arise from noisy or missing open-vocabulary detections.
When either the target or key context objects are missed (or spuriously detected), downstream mapping and verification become unreliable.
A related issue is limited observability: some context objects required by the description are not visible from the viewpoints encountered
before the episode ends, preventing the pipeline from accumulating sufficient evidence for verification.

\noindent\textbf{Ambiguity.}
Some episodes are intrinsically ambiguous: multiple same-category instances remain consistent with the text description under partial
observations, and the caption alone does not provide enough constraints to uniquely identify the intended instance.
In such cases, additional sensing actions (e.g., deliberate viewpoint changes) or external interaction would be required to guarantee success.

\section{Qualitative Comparison with RL-Trained and Training-Free Baselines}
\begin{figure*}[!t]
\subsection{CoIN-Bench}
    \centering
    \includegraphics[width=\linewidth]
    {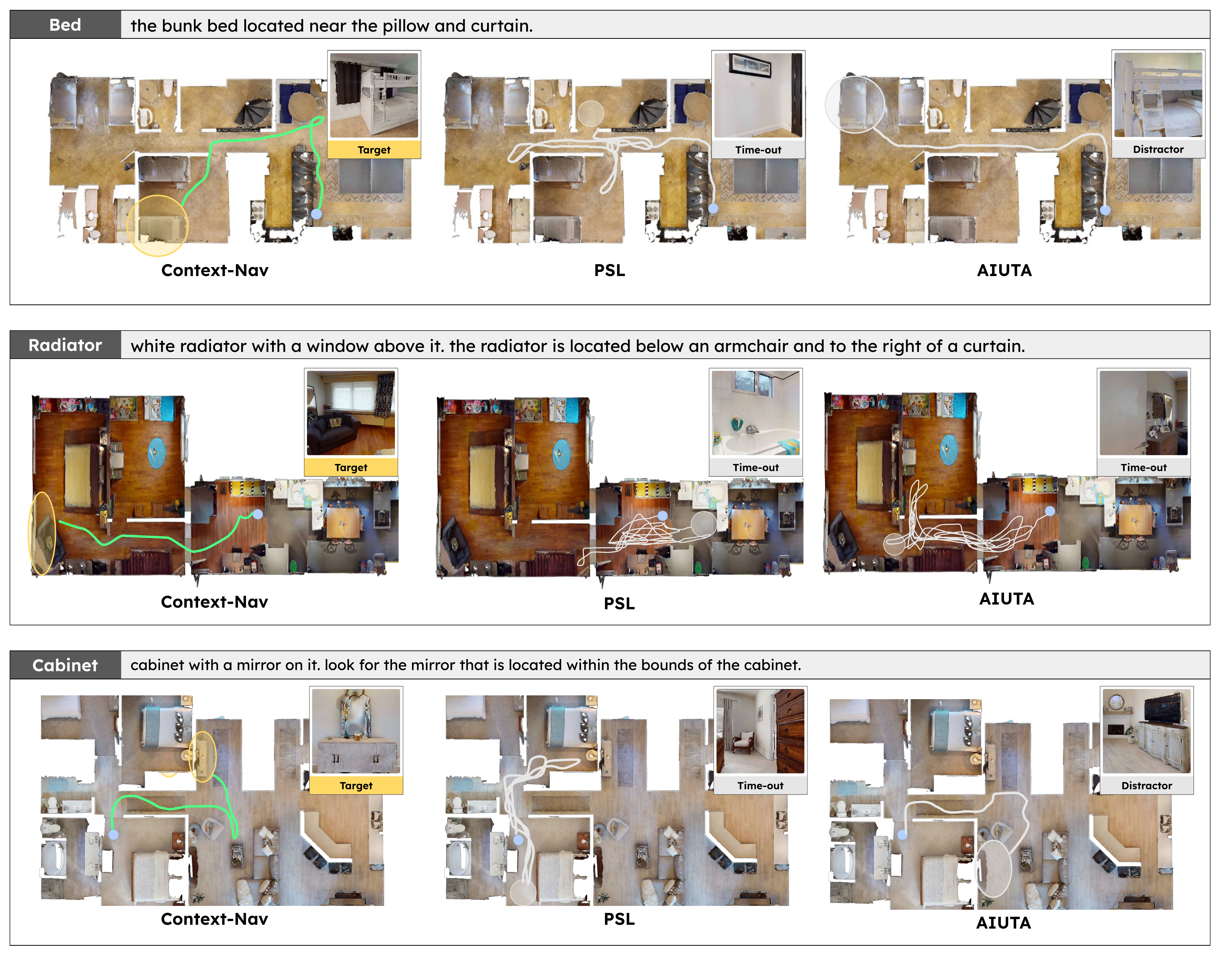}
    \caption{\textbf{Qualitative comparison on CoIN-Bench.} Context-Nav trajectories are compared with those of the RL-trained PSL~\cite{sun2024prioritizedsemanticlearningzeroshot} policy and the training-free AIUTA~\cite{taioli2025coin} agent on CoIN-Bench episodes featuring multiple same-category distractors. For each text goal, top-down trajectories are overlaid on the floor map: Context-Nav is shown in orange and the baselines in light gray. Insets show the final egocentric view and outcome label for each method: \textit{Target} indicates that the correct instance is reached within the step budget, \textit{Time-out} denotes failure to reach any candidate in time, and \textit{Distractor} indicates that the agent stops at a different instance of the same category. In these examples, Context-Nav consistently reaches the correct instance, whereas the baselines either time out or stop at distractors.}
    \label{fig:CoIN_Compare}
\end{figure*}

\begin{figure*}[!t]
\subsection{InstanceNav}
    \centering
    \includegraphics[width=\linewidth]
    {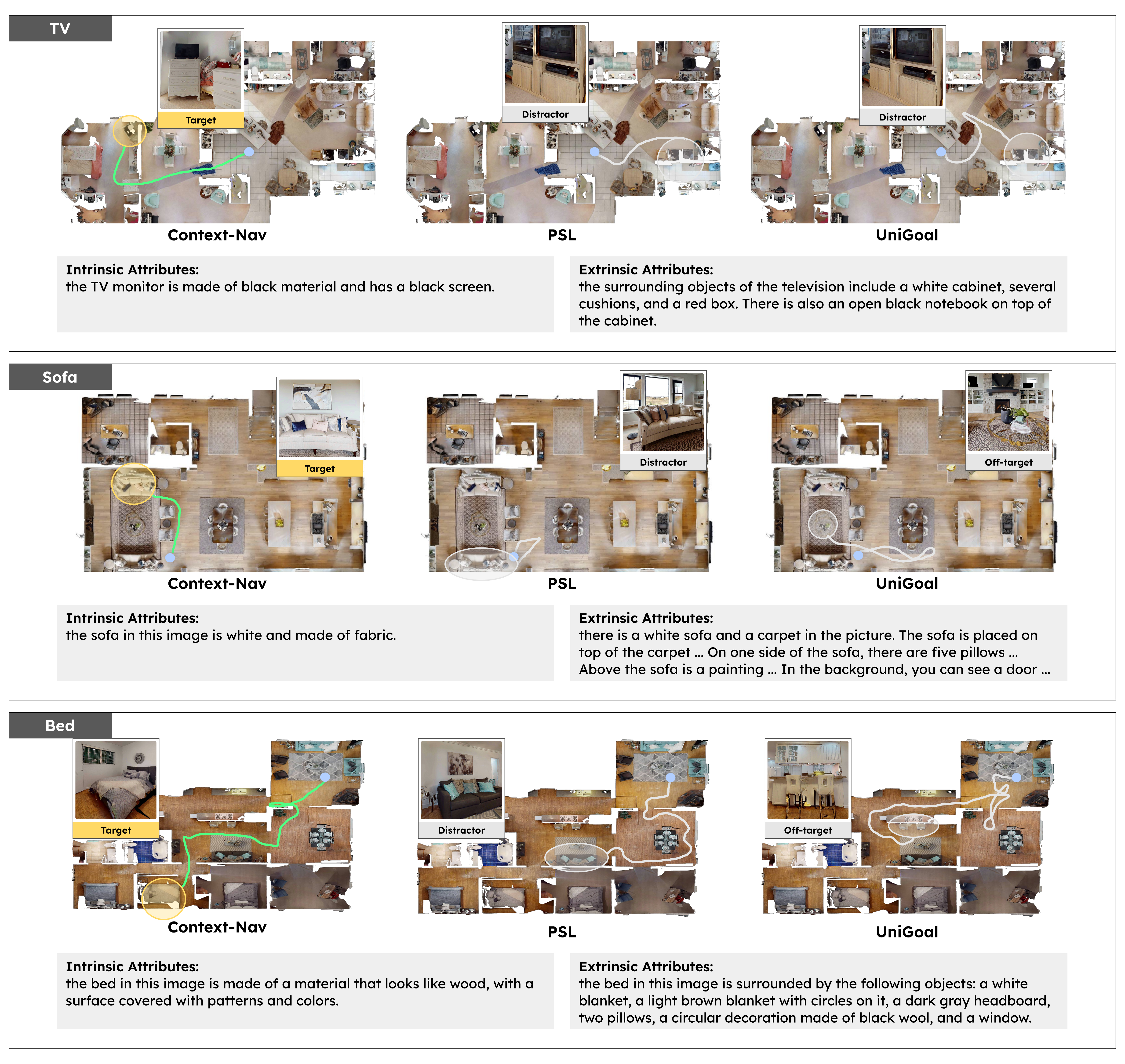}
    \caption{\textbf{Qualitative comparison on InstanceNav.} Representative episodes from InstanceNav~\cite{sun2024prioritizedsemanticlearningzeroshot} compare Context-Nav with the RL-trained PSL~\cite{sun2024prioritizedsemanticlearningzeroshot} policy and the training-free UniGoal~\cite{yin2025unigoaluniversalzeroshotgoaloriented} pipeline. As in Fig.~\ref{fig:CoIN_Compare}, top-down trajectories and final goal views are visualized, drawing Context-Nav in orange and the baselines in light gray. The terminal state is annotated as \textit{Target} when the correct instance is reached, \textit{Distractor} when the agent stops at a different instance of the same category, and \textit{Off-target} when the agent stops at an object from a different category than the described target. These examples highlight that Context-Nav successfully reaches the correct goal instances, whereas PSL and UniGoal often time out, stop at same-category distractors, or end up off-target at objects of incorrect categories.}
    \label{fig:PSL_Compare}
\end{figure*}

Qualitative comparisons with RL-trained and training-free baselines on CoIN-Bench~\cite{taioli2025coin} and InstanceNav~\cite{sun2024prioritizedsemanticlearningzeroshot} complement the quantitative results by visualizing how agents behave in cluttered scenes (Figs.~\ref{fig:CoIN_Compare} and \ref{fig:PSL_Compare}). For each dataset, the RL-trained PSL \cite{sun2024prioritizedsemanticlearningzeroshot} policy serves as the training-based baseline, and a representative training-free pipeline is chosen (AIUTA \cite{taioli2025coin} on CoIN-Bench and UniGoal \cite{yin2025unigoaluniversalzeroshotgoaloriented} on InstanceNav). Each panel visualizes a top-down map together with the natural-language goal, a crop of the target object, and the navigation trajectories of all methods. The trajectory of Context-Nav is drawn in orange, while trajectories of the baseline methods are shown in light gray. The final stopping location of each agent is annotated as \textit{Target} when it reaches the correct instance, \textit{Time-out} when the maximum step budget is exceeded without reaching the goal, \textit{Distractor} when the agent stops at a different instance of the same category, and \textit{Off-target} when it stops at an object from a different category. Across both benchmarks, the illustrated episodes show that Context-Nav more reliably reaches the correct target and tends to follow shorter, context-consistent routes, whereas the baselines often time out or terminate at distractors or off-target objects.

\clearpage

\clearpage

{
    \small
    \bibliographystyle{ieeenat_fullname}
    \bibliography{supp}

}

%% file: preamble.tex
\usepackage{threeparttable}
\usepackage{url}
\usepackage{booktabs}
\usepackage{makecell}
\usepackage{xcolor}
\usepackage{pifont}
\usepackage{multirow}
\usepackage{array}
\usepackage{graphicx}
\usepackage{ulem}
\newcommand{\cmark}{\textcolor{green!60!black}{\ding{51}}} % ✓
\newcommand{\xmark}{\textcolor{red!80!black}{\ding{55}}}
\usepackage{amsmath}
\usepackage{enumitem}
\usepackage[most]{tcolorbox} 
\usepackage{xcolor}
\tcbuselibrary{breakable}

%% file: bodies/00_Abstract.tex
\begin{abstract}
Text-goal instance navigation (TGIN) asks an agent to resolve a single, free-form description into actions that reach the correct object instance among same-category distractors. We present \textit{Context-Nav}, which elevates long, contextual captions from a local matching cue to a global exploration prior and verifies candidates through 3D spatial reasoning. First, we compute dense text-image alignments for a value map that ranks frontiers---guiding exploration toward regions consistent with the entire description rather than early detections. Second, upon observing a candidate, we perform a viewpoint-aware relation check: the agent samples plausible observer poses, aligns local frames, and accepts a target only if the spatial relations can be satisfied from at least one viewpoint. The pipeline requires no task-specific training or fine-tuning; we attain state-of-the-art performance on InstanceNav and CoIN-Bench. Ablations show that (i) encoding full captions into the value map avoids wasted motion and (ii) explicit, viewpoint-aware 3D verification prevents semantically plausible but incorrect stops. This suggests that geometry-grounded spatial reasoning is a scalable alternative to heavy policy training or human-in-the-loop interaction for fine-grained instance disambiguation in cluttered 3D scenes.
\end{abstract}

%% file: bodies/01_introduction.tex
\section{Introduction}
The next frontier in embodied AI hinges on agents that can understand and act upon natural language instructions in real-world 3D environments. Given a free-form description like \textit{``the blue ceramic mug next to the coffee maker, in the kitchen"}, agents should be able to locate a specific instance of an object in an unexplored area. The challenge is not merely object detection, but rather the ability to ground entire descriptions---not just object categories---into coherent spatial reasoning (identifying which blue mug is the correct one). In the process, this text-goal instance navigation (TGIN) requires integration of visual perception, linguistic understanding, and spatial reasoning into a unified decision-making process.

Contemporary methods fall into three broad categories, each with distinct failure modes. First, supervised learning-based approaches \cite{majumdar2023zsonzeroshotobjectgoalnavigation, sun2024prioritizedsemanticlearningzeroshot,khanna2024goatbenchbenchmarkmultimodallifelong} rely on reinforcement learning and semantic supervision; however, they are data-hungry and remain brittle to distribution shifts---failing to generalize to unseen object configurations or novel linguistic descriptions. Next, a zero-shot modular approach \cite{yin2025unigoaluniversalzeroshotgoaloriented} decomposes the problem into scene understanding, online attribute matching, and frontier-based exploration, but matching operations are inherently viewpoint-biased and do not gracefully handle fine-grained spatial constraints buried in the given descriptions. Lastly, an interactive approach \cite{taioli2025coin} resolves ambiguity by asking questions when uncertain; yet, this human-in-the-loop assumption is unrealistic for a variety of deployment scenarios and hardly exploits the context already encoded in the given description. Across these categories, a recurring limitation is the underutilization of the description: most systems reduce longer descriptions to either a set of object labels or a structured representation. Nevertheless, the detailed surrounding environment is not extraneous information but rather a powerful constraint that dramatically narrows the search space and disambiguates among same-category distractors.

To bridge the gap and fully leverage all intrinsic and extrinsic cues in long natural-language descriptions, we design Context‑Nav from a fundamentally different perspective (Fig. 1): spatial reasoning is not a verification step but a primary exploration signal. Rather than detecting objects and then checking whether they match the description, agents should explore spaces that are contextually consistent with the entire description, and only commit to an instance after explicit 3D spatial verification. This shift requires two key innovations. First, we drive exploration---not by committing to early object detections, but by identifying frontier regions that are most consistent with the entire caption. For this, we compute dense text–image alignments in each observation and aggregate them into a global value map---turning long, contextual captions into map‑level exploration signals. Then, frontiers that are most consistent with the caption are naturally prioritized.

Second, we ground candidate object verification in 3D geometry and disambiguate among same-category distractors using spatial relations (e.g., ``left of," ``near," ``in front of")---a geometric challenge due to the inherent viewpoint-dependency, which is often overlooked by existing verification methods. Specifically, we sample viewpoints around the object of interest, align reference frames per viewpoint, and examine whether spatial relations from the caption can be satisfied from any single pose. If not, the candidate is rejected, and exploration resumes. Crucially, this verification process is training-free and uses only the 3D geometry captured by the agent's online map and the room-level spatial context; the agent builds a wall‑only layer to segment rooms.

We comprehensively evaluate Context-Nav on two complementary TGIN benchmarks within HM3D \cite{yadav2023habitatmatterport3dsemanticsdataset}: InstanceNav \cite{sun2024prioritizedsemanticlearningzeroshot} provides separate intrinsic and extrinsic attributes per goal and may include same-category distractors, and CoIN-Bench \cite{taioli2025coin} guarantees multiple distractors and compresses attributes into short descriptions originally designed for interactive agents. Our method achieves state-of-the-art Success Rate (SR) and Success weighted by Path Length (SPL). Furthermore, ablation studies confirm that both context-driven frontier selection and viewpoint-aware relation verification contribute substantially to performance, and qualitative analysis reveals that spatial context---when properly leveraged---disambiguates distractors more reliably than learned similarity functions. These results underscore a principle for embodied AI: spatial reasoning grounded in 3D geometry is a scalable, generalizable alternative to learned policies and human supervision for resolving fine-grained instance ambiguity.

In summary, our contributions are as follows:
\begin{itemize}
\item \textbf{Context‑driven exploration:} We encode long, contextual descriptions into a dense value map for frontier selection---guiding toward semantically relevant regions without premature commitments.
\item \textbf{Viewpoint‑aware 3D relation verification:} We establish a principled spatial reasoning framework that examines relation predicates to accept or reject targets under viewpoint uncertainty.
\item \textbf{No task-specific training:} Our pipeline requires neither TGIN-specific policy training nor fine-tuning---enabling direct transfer to novel scene configurations and open-vocabulary object categories.
\item \textbf{State‑of‑the‑art performance:} Our method consistently achieves superior SR and SPL on InstanceNav and CoIN-Bench---demonstrating its effectiveness across challenging scenarios.
\end{itemize}

%% file: bodies/02_related_work.tex
\begin{figure*}[!t]
    \centering
    \includegraphics[width=\linewidth]
    {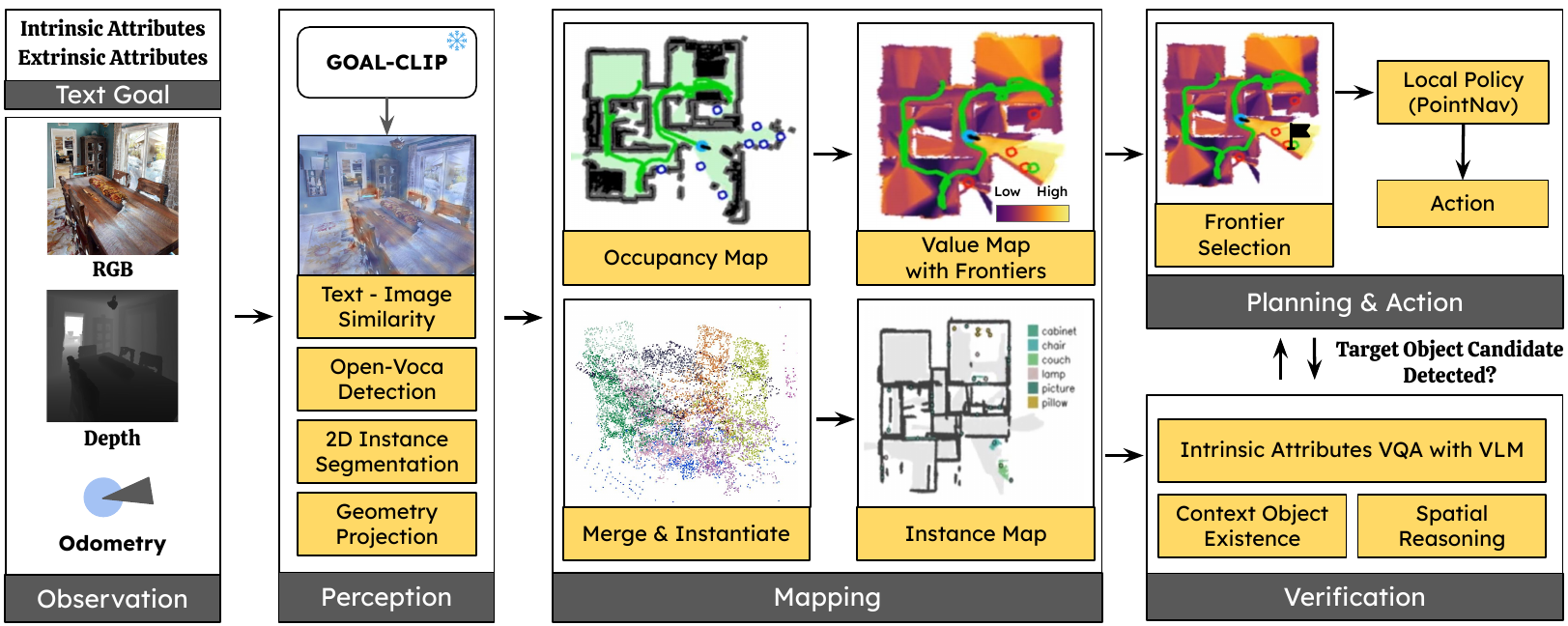}
    \caption{\textbf{Overall pipeline of Context-Nav.} Given RGB-D observations, odometry, and a free-form text goal, the perception and mapping modules use GOAL-CLIP, open-vocabulary detection, and 3D projection to build an occupancy map, a context-conditioned value map, and an instance-level map. Whenever a target object candidate is detected, the verification module checks intrinsic attributes with a VLM and extrinsic attributes through 3D spatial reasoning to decide whether to terminate or continue exploring.}
    \label{fig:methodology}
\end{figure*}

\section{Related Work}
In this section, we highlight key prior work; for a more detailed literature survey, please refer to the supplementary material (Supp. Sec.~B).

\subsection{Visual Navigation}
Visual navigation is a fundamental capability for embodied agents that operate across diverse tasks and environments \cite{Batra2020RearrangementAC, Wang2022LearningLO, Li2022BEHAVIOR1KAB, Srivastava2021BEHAVIORBF}. Visual navigation approaches can be broadly organized along two axes: (i) how the goal is specified (e.g., as metric coordinates, a goal image, a semantic category, or a natural-language description) and (ii) what representation is used for decision-making, ranging from purely geometric maps to semantically enriched scene representations \cite{Savva2019HabitatAP, Wijmans2019DDPPOLN, Batra2020ObjectNavRO, Zhu2016TargetdrivenVN}.

Building on these formulations, end-to-end policies trained with reinforcement learning or imitation learning on proxy tasks achieve strong object navigation performance within a closed label set \cite{ye2021auxiliarytasksexplorationenable, chang2020semantic, deitke2022procthorlargescaleembodiedai, gireesh2022objectgoalnavigationusing, ramrakhya2023pirlnavpretrainingimitationrl}. Zero-shot extensions relax the fixed taxonomy \cite{majumdar2023zsonzeroshotobjectgoalnavigation, alhalah2022zeroexperiencerequiredplug, sun2024prioritizedsemanticlearningzeroshot} and training-free modular pipelines decompose perception, mapping, and planning \cite{gadre2022cowspasturebaselinesbenchmarks, yokoyama2023vlfmvisionlanguagefrontiermaps, zhou2023escexplorationsoftcommonsense, Yu_2023, wu2024voronavvoronoibasedzeroshotobject, kuang2024openfmnavopensetzeroshotobject, zhang2024trihelperzeroshotobjectnavigation, yin2024sgnavonline3dscene, cai2023bridgingzeroshotobjectnavigation, long2024instructnavzeroshotgenericinstruction}. However, most of these systems ultimately ground language at the category level or via short attribute snippets; they treat detection-driven matches as the primary exploration signal. % , and bias frontier-based exploration using vision-language signals: frontiers are ranked via CLIP similarities or text-conditioned value maps, and large language or vision-language models reason over detected-object lists or semantic maps to score candidate frontiers 

\subsection{Instance-level navigation}
Instance-level navigation aims to select the correct object instance among same-category distractors under open-vocabulary descriptions, rather than merely reaching any instance of a category. Personalized or user-linked navigation ties goals to specific physical objects \cite{barsellotti2025personalizedinstancebasednavigationuserspecific, dai2024thinkactaskopenworld}, while open-vocabulary ObjectNav benchmarks expand label coverage yet typically retain category-level success metrics \cite{yokoyama2024hm3dovondatasetbenchmarkopenvocabulary}. Training-based text-goal instance navigation (TGIN) methods seek to bridge text-image granularity by carefully designed curricula or inference-time goal expansion \cite{sun2024prioritizedsemanticlearningzeroshot}, and interaction-centric agents solicit additional attributes from users when the goal is ambiguous \cite{taioli2025coin}. Despite this progress, open-vocabulary spatial reasoning and fine-grained attribute recognition remain fragile under partial observations or long-range views \cite{yang2025thinking, bianchi2024devilfinegraineddetailsevaluating, bravo2023openvocabularyattributedetection}, and relation handling in prior modular systems is typically restricted to category-level cues or viewpoint-agnostic heuristics \cite{yin2025unigoaluniversalzeroshotgoaloriented}.

Our work focuses on TGIN from an essentially different angle: long, contextual captions are not only a verification signal, but can also guide semantic exploration. Furthermore, we design viewpoint-aware 3D relation reasoning to resolve uncertainty, while remaining training-free and avoiding human-in-the-loop interaction. 

%% file: bodies/03_method.tex
\section{Method}
In this section, we describe Context-Nav  for context-driven exploration and 3D spatial reasoning. Fig. \ref{fig:methodology} illustrates the overall pipeline.

\subsection{Problem formulation}
At time step \textit{t}, the agent observes an RGB image $X_t \in \mathbb{R}^{H \times W \times 3}$ and a depth map $D_t \in \mathbb{R}^{H \times W}$ and estimates its egocentric pose $T_t \in SE(3)$. Given a free-form natural language goal $G$ containing intrinsic attributes (e.g., color, shape, material of the target) and extrinsic/context attributes (e.g., named surrounding objects and their spatial relations), the agent executes a sequence of actions $\mathcal{A} =$ \{\texttt{forward, turn-left, turn-right, stop}\} to reach the unique described instance. The environment may contain multiple instances of the same category with different attributes; the agent has no prior map and must disambiguate using only $G$, its sensors, and on‑the‑fly mapping. Open‑vocabulary names are allowed. The map grows incrementally, and many relations (e.g., left of, behind) depend on the narrator’s unknown viewpoint.

\subsection{Perception and Online 3D Mapping}
\noindent\textbf{Open-Vocabulary Detection and Verification.} We generate landmark and observation clouds through three steps: 1) category canonicalization, 2) open‑vocabulary detection, and 3) VLM verification. First, we parse noun phrases from $G$ and map them to a canonical category set $\mathcal{C}$ (e.g., $couch \rightarrow  sofa$) to form the prompt list $\mathcal{C}_G$ (please refer to Supp. Sec. C for $G$ preprocessing). For every new observation, we run an open‑vocabulary detector~\cite{liu2024groundingdinomarryingdino} to obtain bounding boxes ${b_i}$. For each $b_i$, we derive a pixel mask $m_i$ using a segmenter~\cite{zhang2023fastersegmentanythinglightweight}. Then, verification branches by category membership. If the proposed class lies within the COCO taxonomy, we confirm the prediction with a class‑specific detector~\cite{wang2022yolov7trainablebagoffreebiessets} and accept high‑confidence detections without further inspection. For open‑set (non‑COCO) categories, we query a VLM with a yes/no question about the outlined region (e.g., “Is the outlined object a \texttt{<category>}?”). Verified instances are fused into a persistent landmark cloud, while unverified proposals are stored in an observation cloud for later re‑association. This design choice---deferring VLM queries to only open-set categories---reflects an optimization for computational efficiency while maintaining confidence.

\noindent\textbf{Instance-Level 3D Mapping.} 
As the agent explores, we associate observations from multiple viewpoints into per-instance 3D point clouds. We decouple the association process into two passes for computational efficiency.
\begin{itemize}
    \item \textit{Spatial proximity heuristic:} For candidate and reference instance point clouds, we first check if their 2D centers (projected into 3D using $D_t$ and $T_t$) are sufficiently close. If so, we assume short-term tracking consistency (the agent revisits nearby regions within a short time window) and merge immediately---avoiding expensive per-point comparisons.
    \item \textit{Voxel-overlap verification:} If the proximity heuristic fails, we uniformly sub‑sample and discretize both point clouds into voxel grids at a fixed resolution and compare them using voxel intersection normalized by the smaller set:
    \begin{equation}
        s(A,B)=\frac{|S_A\cap S_B|}{\min(|S_A|,|S_B|)},
    \label{equ:overlap}
    \end{equation}
    \noindent where $S_A$ and $S_B$ are the respective voxel sets. If $s(A, B)$ exceeds a threshold, the two observations are merged into the same instance; otherwise, a new instance is created~\cite{yang2023sam3dsegment3dscenes}.
\end{itemize}

\noindent\textbf{Wall-Only Map.} 
For meticulous spatial reasoning, we maintain a separate wall map derived solely from structural planes; the occupancy map includes furniture and clutter, which can spuriously fragment free space into disconnected regions. To build a wall-only map, we range-gate and height-gate (0.8-3 m) depth points, and iteratively segment multiple vertical planes using RANSAC~\cite{Zhou2018, Fischler1981RANSAC}. Planes with insufficient height are discarded; the union of accepted inliers yields a wall point cloud. Rasterizing this cloud produces a binary wall layer. The connected component analysis of the free space in this layer defines rooms; two objects are considered to be in the same room if their centroids fall within the same component. Line-of-sight checks are gated by ray-casting through the wall layer---ensuring that non-structural obstacles do not cause erroneous fragmentation.
 
\subsection{Context-Driven Exploration}
\noindent\textbf{Text-Conditioned Value Map and Frontiers.}
We guide exploration by constructing a context‑conditioned value map. Given the goal $G$, we build a prompt that concatenates the target category with its attributes and context. Standard CLIP struggles with long descriptive prompts; we therefore employ GOAL (Global–Local Object Alignment Learning)~\cite{choi2025goalgloballocalobjectalignment}, a fine-tuning of CLIP that aligns long text with image regions by matching local image–sentence pairs and propagating token-level correspondences. We encode $G$ and each observation $X_t$ with GOAL and compute per‑pixel similarities. Using depth and pose, we project these similarities into a top‑down grid, forming a dense value map $V_t$~\cite{yokoyama2023vlfmvisionlanguagefrontiermaps}. Frontier cells at the boundary between explored free space and unknown space are ranked by their values; the agent chooses the frontier with the highest value to explore next.

\noindent\textbf{Room-Level Constraint.} While value-map-based frontier ranking is generally effective, we introduce a single override mechanism to leverage a room-level constraint: if (i) a target instance has been detected, (ii) at least one context instance remains unobserved in the target's wall-bounded room, and (iii) unexplored frontiers exist within that room, then we override the global highest-value frontier exactly once and select the nearest unexplored frontier within the room (via geodesic distance through explored free space). After this waypoint is reached, the override flag is consumed and the default value-map policy resumes.

\noindent\textbf{Local Policy and Action.} Low‑level motion is handled by an off-the-shelf depth‑only point‑goal navigation policy~\cite{anderson2018evaluationembodiednavigationagents} (Variable Experience Rollout~\cite{wijmans2022verscalingonpolicyrl, yokoyama2023vlfmvisionlanguagefrontiermaps} on HM3D). Given a waypoint from the planner, this policy outputs discrete actions $\mathcal{A}$ until the waypoint is reached.

\subsection{Instance Verification}
Once a candidate instance is detected, we verify both its intrinsic attributes through VQA and extrinsic attributes through grounding in 3D geometry.

\noindent\textbf{Intrinsic Attributes.}
We parse $G$ with an LLM to derive multiple yes/unknown/no question prompts for each attribute: paraphrasing reduces brittleness \cite{taioli2025coin}. Upon observing a candidate instance, we query the VLM using the current RGB frame $X_t$, with the instance highlighted. The VLM outputs a scalar confidence score $s\in\{0,\dots,15\}$, discretized into three bins:
\begin{itemize}
    \item \textit{No} (0-4): Attribute clearly absent.
    \item \textit{Unknown} (5-10): Unclear from current view.
    \item \textit{Yes} (11-15): Attribute clearly present.
\end{itemize}
An attribute is satisfied if any of its prompts returns \textit{Yes}. If all prompts are \textit{No} even after re‑querying, the instance fails. To handle unknown results caused by viewpoint ambiguity, we defer judgment and re-query using the most informative frame from the next five steps. Specifically, we log five RGB frames, score each by text–image similarity, and re-issue the attribute question on the highest-scoring frame. This adaptive re-querying handles viewpoint-dependent ambiguity (e.g., color may be unclear in shadow but evident in direct light) without requiring active object manipulation.

\noindent\textbf{Extrinsic Attributes.}
Our innovative, explicit 3D, viewpoint-aware verification of extrinsic attributes consists of four steps. Please refer to Supp. Sec. D for an illustrated overview and additional discussion.

\noindent\emph{Step 1. Room-Level Filtering.} We first ensure that the target instance and at least one context instance are co-located in the same wall-bounded room (within 3 m via geodesic distance); this filters out spurious relation checks. Let \((\mathrm{ref},\mathrm{tgt},\rho)\) denote a spatial-relation triple parsed from $G$, where ref and tgt are object labels and $\rho$ is the relation type. From each triple, we generate candidate instance pairs, the effective relation set $\mathcal{T}$, and the center set $\mathcal{S}=\{c_i (x_i,y_i)\}$.

\noindent\emph{Step 2. Candidate Viewpoint Sampling.} To handle viewpoint ambiguity, we define a candidate set $\mathcal{V}$ as follows. For an anchor $m$ (either the relation-pair midpoint or the global centroid of $\mathcal{S}$), we place $N_\theta{=}24$ candidate viewpoints at evenly spaced bearings, with radius $r\in\{0.8,1.2,1.6,2.0\}$:
\begin{equation}
v_{r,k}(m)=m+r\,(\cos\theta_k,\ \sin\theta_k)^\top,
\end{equation}
where $\theta_k{=}2\pi k/24, k=0,\dots,23$. The union over anchors, radii, and angles defines $\mathcal{V}$.

\noindent\emph{Step 3. Viewpoint Alignment.} For each candidate $v\in\mathcal{V}$ and a triple with reference center $c_r$ and target center $c_t$, we construct a viewpoint-aligned local frame at $v$ so that $+\hat x$ points from $v$ to $c_r$. We define yaw from $v$ to $c_r$ as
\begin{equation}
\psi=\mathrm{atan2}\big((c_r)_y-v_y,\ (c_r)_x-v_x\big),
\end{equation}
and unit axes as $u_x=(\cos\psi,\ \sin\psi)$ and $u_y=(-\sin\psi,\ \cos\psi)$. Any point $q$ is transformed to viewpoint-aligned coordinates via inner products: 
\begin{equation}
\tilde x(q)=\langle q{-}v,\,u_x\rangle,\qquad \tilde y(q)=\langle q{-}v,\,u_y\rangle,
\end{equation}
where $\langle\cdot,\cdot\rangle$ is the Euclidean inner product in the ground plane; the bearing in this frame is $b(x,y){:=}\mathrm{atan2}(y,x)$ and $b(\tilde r){=}0$ for the aligned reference $\tilde r$, where $\tilde\cdot$ denotes quantities in the viewpoint-aligned frame.

\noindent\emph{Step 4. Relation Predicates and Validation.} We define binary predicates for seven spatial relations. By letting tolerances be $\varepsilon_m{=}0.15$\,m (position), $\varepsilon_\theta{=}25^\circ$ (bearing), $d_{\mathrm{near}}{=}2.0$\,m (near), and $\varepsilon_z{=}0.15$\,m (height):
\begin{equation}
\label{eq:spatial-preds}
\begin{aligned}
\textbf{left:}&\ \ \tilde y(c_t)-\tilde y(c_r)\ \ge\ \varepsilon_m,\\
\textbf{right:}&\ \ \tilde y(c_r)-\tilde y(c_t)\ \ge\ \varepsilon_m,\\
\textbf{front:}&\ \ |b(\tilde t)|\ \le\ \varepsilon_\theta\ \land\ \tilde x(c_t)\ \le\ \tilde x(c_r)-\varepsilon_m,\\
\textbf{behind:}&\ \ |b(\tilde t)|\ \le\ \varepsilon_\theta\ \land\ \tilde x(c_t)\ \ge\ \tilde x(c_r)+\varepsilon_m,\\
\textbf{near:}&\ \ \|c_t-c_r\|_2\ \le\ d_{\mathrm{near}},\\
\textbf{above:}&\ \ \hat{z}_t-\hat{z}_r\ \ge\ \varepsilon_z, \\
\textbf{below:}&\ \ \hat z_r-\hat z_t\ \ge\ \varepsilon_z\quad
\end{aligned}
\end{equation}
where $\hat z$ denotes per-instance height estimates from fused point clouds.

For a candidate to be accepted, there should exist at least one viewpoint $v^* \in \mathcal{V}$ such that: 1) $v^*$ lies in the same wall-bounded room as both endpoints of every relation triple in $\mathcal{T}$ and 2) all spatial predicates corresponding to the relation triples are satisfied simultaneously at $v^*$. If such a viewpoint exists, the target is confirmed; otherwise, exploration resumes and the instance is discarded.

%% file: bodies/04_experiments.tex
\section{Experiments}
\begin{table*}[!t]
    \centering
    \input{tables/main_experiment} 
    \label{tab:Quantitaive_result}
    \caption{\textbf{Benchmark results on InstanceNav and CoIN-Bench}. Comparison of RL-trained policies, training-free modular baselines, and the proposed Context-Nav on InstanceNav and the three CoIN-Bench splits (\texttt{Val Seen}, \texttt{Val Seen Synonyms}, \texttt{Val Unseen}). Input type c denotes a category-level goal specification, while d denotes a language description of the target.}
\end{table*}

\subsection{Settings}
\noindent\textbf{Datasets.}
We evaluate performance on two TGIN benchmarks in HM3D \cite{ramakrishnan2021hm3d} via Habitat \cite{savva2019habitatplatformembodiedai}: 
\begin{itemize}
    \item \textit{InstanceNav} \cite{sun2024prioritizedsemanticlearningzeroshot} augments Instance Image Navigation (IIN) \cite{krantz2022instancespecificimagegoalnavigation} with goal texts that explicitly separate intrinsic (appearance) and extrinsic (contextual) attributes for instance disambiguation; the test set contains 1{,}000 episodes featuring 795 unique objects across 36 scenes and six categories (chair, sofa, TV, bed, toilet, and plant). 
    \item \textit{CoIN‑Bench}~\cite{taioli2025coin} filters GOAT‑Bench episodes with distractors; the dataset contains 831 \texttt{Val Seen}, 359 \texttt{Val Seen Synonyms}, and 459 \texttt{Val Unseen}, with language goals automatically constructed by combining simulator semantics with BLIP‑2 attributes and then verbalized by ChatGPT‑3.5 (category coverage: Supp. Sec. H).
\end{itemize}

\noindent\textbf{Evaluation metrics.}
We report Success Rate (SR) and Success weighted by Path Length (SPL), which is defined as $\mathrm{SPL}=\tfrac{1}{N}\sum_{i=1}^{N} S_i \cdot \tfrac{\ell_i}{\max(p_i,\ell_i)}$. Success criteria follow the standard per dataset: for InstanceNav, reaching the target within 1m and at most 1{,}000 steps~\cite{yin2025unigoaluniversalzeroshotgoaloriented}; for CoIN‑Bench, reaching the target within 0.25m and at most 500 steps~\cite{taioli2025coin}.

\noindent\textbf{Baselines.}
We compare Context-Nav against both learning-based and training-free baselines: PSL \cite{sun2024prioritizedsemanticlearningzeroshot}, GOAT \cite{khanna2024goatbenchbenchmarkmultimodallifelong}, VLFM \cite{yokoyama2023vlfmvisionlanguagefrontiermaps}, AIUTA \cite{taioli2025coin}, and UniGoal \cite{yin2025unigoaluniversalzeroshotgoaloriented}.

\begin{figure*}[!t]
    \centering
    \includegraphics[width=\linewidth]
    {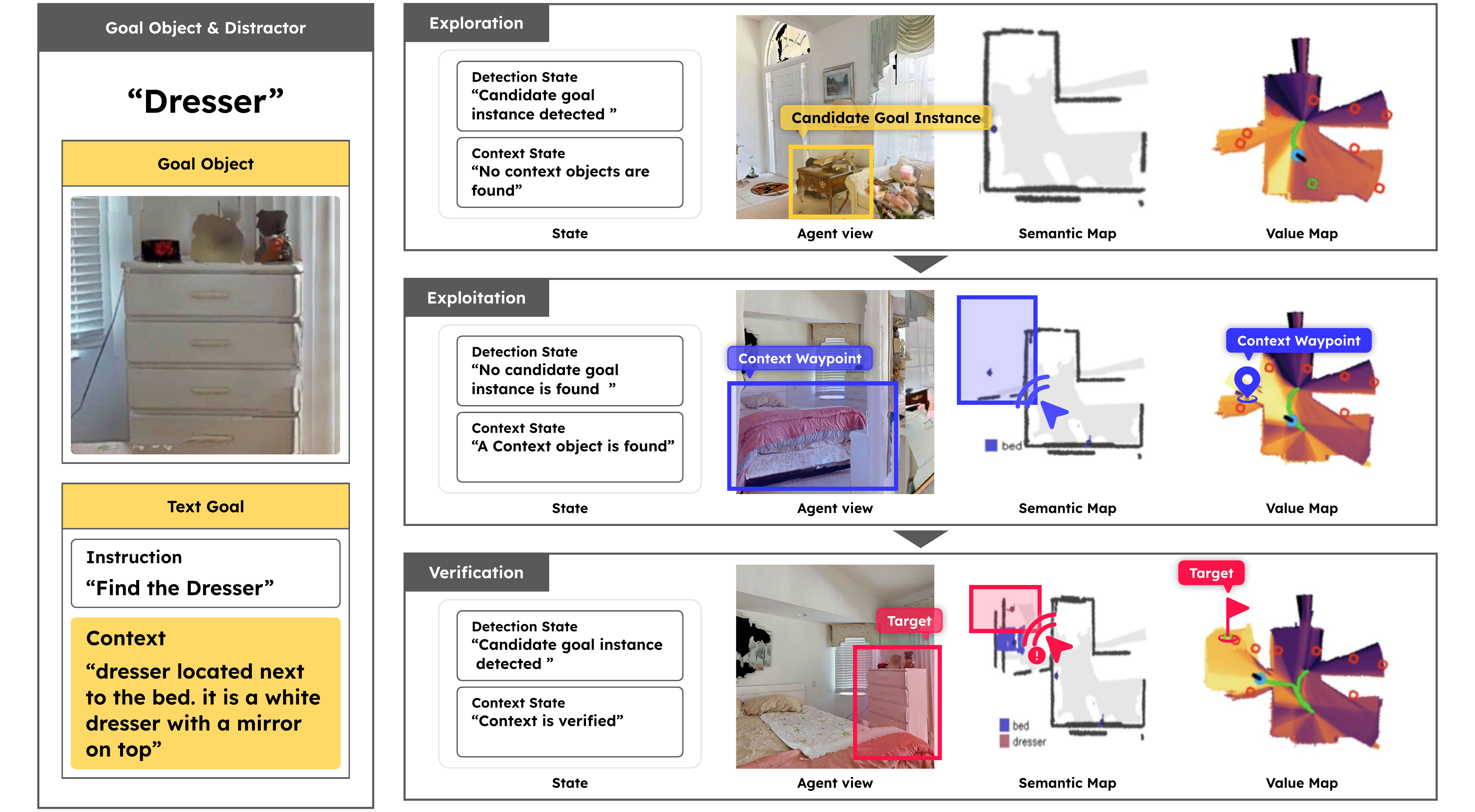}
    \caption{\textbf{Stage-wise qualitative example of context-driven navigation.} An episode where the agent must find a dresser described as “located next to the bed” and “a white dresser with a mirror on top”. Early dresser candidate is not selected because context objects are absent; after the bed is detected, the map concentrates around the corresponding room, frontier selection focuses on that area, and a dresser that satisfies both intrinsic attributes and 3D spatial relations with the bed and mirror is finally verified as the goal.} 

    \label{fig:step_result}
\end{figure*}

\subsection{Benchmark Results}
Table 1 compares Context-Nav against baselines. Across InstanceNav and all CoIN-Bench splits, Context-Nav achieves state-of-the-art SR among both RL-trained and training-free baselines while requiring neither policy training nor human interaction. On InstanceNav, our method attains the highest SR (26.2\%), exceeding the RL-trained PSL policy and outperforming the training-free UniGoal pipeline by a substantial margin. This demonstrates that context-driven exploration and 3D relation verification can close—and even invert—the performance gap with trained policies. On CoIN-Bench, Context-Nav likewise yields the highest SR among all compared methods, including GOAT, VLFM, UniGoal, and the interactive AIUTA baseline, showing that a training-free, open-vocabulary pipeline can outperform both RL and human-in-the-loop alternatives.

\subsection{Ablation Study}
We perform ablations to answer two questions on the CoIN‑Bench Val Seen Synonyms split: 
(i) how much the \textit{language signal} (backbone $\times$ prompt) is beneficial; and
(ii) which \textit{modules} are load‑bearing for instance disambiguation.

\input{tables/ablation}
\begin{table}[!t]
    \centering
    \AbBackbone
    \caption{\textbf{Ablation of similarity backbone and prompt on CoIN-Bench Val Seen Synonyms.} We compare BLIP-2 and GOAL-CLIP under different prompt designs (category only, category with intrinsic attributes, and full contextual text).}
    \label{tab:ablation_backbone}
\end{table}

\noindent\textbf{Similarity Backbone and Prompt.}
Table~\ref{tab:ablation_backbone} shows that the agent benefits most when the alignment model can exploit \textit{full, contextual text}; \textit{GOAL‑CLIP} with the full caption outperforms the same backbone with category‑only by +6.6 SR / +3.3 SPL, and beats category\,+\,intrinsic by +3.6 / +1.2.
For \textit{BLIP‑2}, full captions mainly help efficiency (SPL: +2.2 vs category; +1.3 vs intrinsic) with smaller or no SR gains, suggesting that token‑level grounding (as in GOAL‑CLIP) turns long text into stronger spatial priors than global caption pooling. The overall intuition is that long captions narrow the search to context‑consistent regions before any detection is trusted. Category‑only prompts yield broad, noisy value maps; adding intrinsic cues removes some lookalikes; adding \textit{context} completes the disambiguation signal that our verification later checks in 3D.

\begin{table}[!t]
    \centering
    \AbMethod
    \caption{\textbf{Ablation of pipeline components on CoIN-Bench Val Seen Synonyms.} Replacing value-map–guided frontier ranking with a nearest-frontier heuristic or removing VLM category, attribute, or context verification each degrades SR and SPL.}
    \label{tab:ablation_method}
\end{table}

\noindent\textbf{Effect of Each Pipeline Design.} 
Table \ref{tab:ablation_method} analyzes the contribution of each component as follows:
\begin{itemize}
    \item \textit{Nearest‑frontier instead of value‑map ranking} loses -9.7 SR / -6.3 SPL: without context-driven exploration, the agent detours into wrong rooms and burns steps in semantically irrelevant regions.
    \item \textit{No VLM category verification} for open‑set classes costs -9.2 / -3.8:  open‑set categories drift toward frequent COCO labels---exposing detector bias and open‑vocabulary blind spots.
    \item \textit{No intrinsic attribute checks} degrade -7.8 / -3.2: visually similar same‑category instances (e.g., color/shape variants) get mistakenly accepted.   
    \item \textit{No context (relation) verification} impairs -8.3 / -2.5: the agent stops at the correct object type but in the wrong neighborhood or relation.
\end{itemize}

\subsection{Qualitative Results}
Fig.~\ref{fig:step_result} illustrates typical episodes decomposed into exploration, exploitation, and verification stages. In the early exploration phase, the value map highlights regions that are loosely consistent with the caption; however, the agent does not commit because the context objects (bed or mirror) are absent---thus, no 3D relation validation occurs. As exploration progresses and context instances are detected, the value map sharpens around the corresponding room, making frontier selection more effective. Eventually, a candidate instance satisfying both intrinsic attributes and spatial relations is verified, and the agent stops. Additional examples across diverse target categories demonstrate consistent behavior: the agent seeks semantically relevant rooms and furniture groups, rather than greedily chasing detections. 

\begin{figure*}[!t]
    \centering
    \includegraphics[width=\linewidth]
    {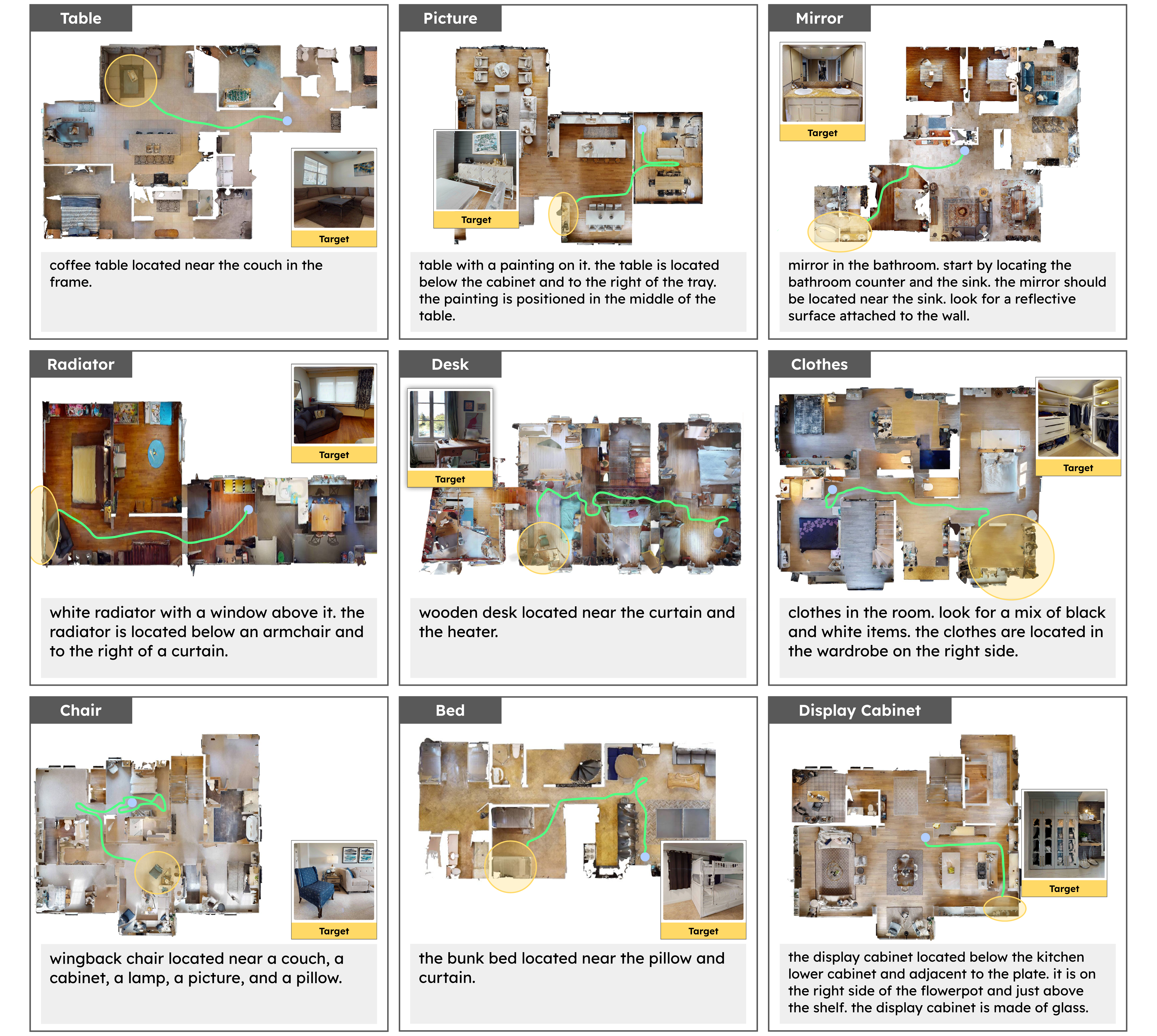}
    \caption{\textbf{Qualitative results across diverse categories and context descriptions.} Successful episodes on CoIN-Bench for nine different target categories, showing top-down trajectories and corresponding goal views. The instructions span a wide range of natural language, from captions that only specify extrinsic context to descriptions that combine intrinsic and extrinsic attributes, and from short hints to detailed multi-sentence goals.}

    \label{fig:Qulatitive}
\end{figure*}

Fig. \ref{fig:Qulatitive} presents additional successful CoIN‑Bench trajectories across nine target categories. The instructions span a wide spectrum of natural language, ranging from purely extrinsic cues (surrounding objects and relations) to captions that mix intrinsic and extrinsic attributes, and from brief hints to multi-sentence descriptions. Across these cases, Context-Nav converts the full description into a value map prior and enforces 3D spatial consistency---steering the agent toward semantically relevant rooms and furniture groupings rather than chasing isolated detections. Additional qualitative comparisons and failure case analyses are included in the supplementary material (Supp. Sec. I-J).

%% file: tables/main_experiment.tex
\centering
\begin{threeparttable}

\small
\begin{tabular}{l cc cc cc cc cc}
\toprule
\multirow{4}{*}{Method} 
& \multicolumn{2}{c}{Model Condition} 
& \multicolumn{2}{c}{InstanceNav} 
& \multicolumn{6}{c}{CoIN-Bench} \\
\cmidrule(lr){2-3}\cmidrule(lr){4-5}\cmidrule(lr){6-11}
 & \multirow{2}{*}{Input} & \multirow{2}{*}{Training-free}
 & \multirow{2}{*}{SR$\uparrow$} & \multirow{2}{*}{SPL$\uparrow$}
 & \multicolumn{2}{c}{Val Seen} 
 & \multicolumn{2}{c}{Val Seen Synonyms} 
 & \multicolumn{2}{c}{Val Unseen} \\
\cmidrule(lr){6-7}\cmidrule(lr){8-9}\cmidrule(lr){10-11}
 &  &  &  &  
 & SR$\uparrow$ & SPL$\uparrow$
 & SR$\uparrow$ & SPL$\uparrow$
 & SR$\uparrow$ & SPL$\uparrow$ \\
\midrule
GOAT \cite{khanna2024goatbenchbenchmarkmultimodallifelong}  & d & \xmark 
& 17.0 & 8.8 & 6.6 & 3.1  
& 13.1 & 6.5 
& 0.2 & 0.1 \\
PSL\tnote{$\dagger$} \cite{sun2024prioritizedsemanticlearningzeroshot}  & d & \xmark 
& 26.0 & 10.2
& 8.8 & 3.3
& 8.9 & 2.8  
& 4.6 & 1.4 \\
\midrule
VLFM \cite{yokoyama2023vlfmvisionlanguagefrontiermaps}  & c & \cmark 
& 14.9 & 9.3  
& 0.4 & 0.3  
& 0.0 & 0.0 
& 0.0 & 0.0 \\
AIUTA \cite{taioli2025coin} & c & \cmark 
& - & -
& 7.4 & 2.9
& 14.4 & 8.0
& 6.7 & 2.3\\

UniGoal \cite{yin2025unigoaluniversalzeroshotgoaloriented}  & d & \cmark 
& 20.2 & \textbf{11.4}  
& 2.8 & 2.4 
& 3.9 & 3.2
& 2.6 & 2.2 \\

\textbf{Ours} & d & \cmark 
& \textbf{26.2} & 9.1
& \textbf{13.5} & \textbf{6.7} 
& \textbf{20.3} & \textbf{10.9}
& \textbf{11.3} & \textbf{5.2} \\
\bottomrule
\end{tabular}
\begin{tablenotes}[flushleft]
\footnotesize
\item[$^\dagger$] Correction for PSL \cite{sun2024prioritizedsemanticlearningzeroshot} whose performance was reported as SR=16.5 and SPL=7.5 \cite{yin2025unigoaluniversalzeroshotgoaloriented}; we have re-run the official code following the standard setting (success distance 1m and 1,000 max steps per episode) and reproduced the performance.
\end{tablenotes}
\label{table:main_result}
\end{threeparttable}

%% file: tables/ablation.tex
\newcommand{\AbBackbone}{
\centering
\begin{tabular}{@{}cccc@{}}
\toprule
\multirow{2}{*}{Backbone}
& \multirow{2}{*}{Prompt}

& \multicolumn{2}{c}{CoIN-Bench} \\
\cmidrule(l){3-4}
  &  & SR$\uparrow$ & SPL$\uparrow$ \\
\midrule
%\multirow{3}{*}{CLIP} & Category &  &  \\   & Intrinsic text & -  & - \\  & Full text & - & - \\
%\midrule
\multirow{3}{*}{BLIP-2 \cite{li2023blip}} & Category & 15.9 & 7.3 \\   & Intrinsic Attributes & 17.8  & 8.2 \\  & Full text & 16.4 & 9.5 \\
\midrule
\multirow{3}{*}{GOAL-CLIP \cite{choi2025goalgloballocalobjectalignment}} & Category & 13.7 & 7.6 \\   & Intrinsic Attributes & 16.7  & 9.7 \\  & \textbf{Full text} & \textbf{20.3} & \textbf{10.9} \\
\bottomrule
\end{tabular}

%\label{tab:ablation}
}

\newcommand{\AbMethod}{
\centering
\begin{tabular}{@{}ccc@{}}
\toprule
\multirow{2}{*}{Method}
& \multicolumn{2}{c}{CoIN-Bench} \\
\cmidrule(l){2-3}
 & SR$\uparrow$ & SPL$\uparrow$ \\
\midrule
Nearest frontier exploration & 10.6  & 4.6 \\
Remove VLM category verification  & 11.1 & 7.1 \\
%Replace wall map to occupncy map & 20.1 & 4.9 \\
Remove attribute verification & 12.5  & 7.7 \\
Remove context verification & 12.0  & 8.4 \\
\midrule
\textbf{Full Approach} & \textbf{20.3}
                 & \textbf{10.9}
               \\
\bottomrule
\end{tabular}
%\label{tab:ablation}
}

%% file: bodies/05_conclusion.tex
\section{Conclusion}
In this work, we presented Context-Nav, a training-free framework for instance-level navigation. Our key design insight is that long, contextual captions can be elevated from a post-hoc verifier to a \textit{primary exploration signal}. Moreover, Context-Nav combines context-driven value maps with viewpoint-aware 3D relation verification, which effectively handles viewpoint uncertainty. We achieved substantial performance gains on InstanceNav and CoIN-Bench without task-specific training---demonstrating that context-driven exploration and viewpoint-aware relation checks enable robust performance. We believe geometry-grounded verification is a scalable complement to learned policies; future work will tighten relation semantics, integrate uncertainty into frontier ranking, and reduce computational latency.

%% file: bodies/06_Acknowledgement.tex
\fontsize{8}{10}\selectfont{%
\noindent\textbf{Acknowledgement.}
This research was partly supported by the Institute of Information \& communications Technology Planning \& Evaluation (IITP) grant funded by the Korea government (MSIT) (No. RS-2022-II220926, Development of Self-directed Visual Intelligence Technology Based on Problem Hypothesis and Self-supervised Methods); by the National Research Foundation of Korea (NRF) grant funded by the Korea government (MSIT) (No. NRF-2022R1C1C1009989); and by the National Research Council of Science \& Technology (NST) grant by the Korea government (MSIT) (No. GTL25041-000).} 